\documentclass[11pt,letterpaper]{article}
\usepackage[letterpaper]{geometry}
\usepackage{acl2012}
\usepackage{times}
\usepackage{latexsym}
\usepackage{amsmath}
\usepackage{multirow}
\usepackage{todonotes}
\usepackage{comment}
\usepackage{amssymb}
\usepackage{flafter}
\usepackage{url}
\makeatletter
\newcommand{\@BIBLABEL}{\@emptybiblabel}
\newcommand{\@emptybiblabel}[1]{}
\makeatother
\usepackage{floatrow}
\usepackage[outercaption]{sidecap} 

\title{Leveraging Sparse and Dense Feature Combinations for Sentiment Classification }

\author{
  Tao Yu \\
  DSI, Columbia University \\
  New York, NY, USA \\
  {\tt ty2326@columbia.edu} \\
  \And
  Christopher Hidey \\
  CS, Columbia University \\
  New York, NY, USA \\
  {\tt ch3085@columbia.edu} \\
  \AND
  Owen Rambow \\
  CCLS, Columbia University  \\
  New York, NY, USA \\
  {\tt rambow@ccls.columbia.edu} \\
  \And
  Kathleen McKeown \\
  CS, Columbia University  \\
  New York, NY, USA \\
  {\tt kathy@cs.columbia.edu} \\}

\date{}

\begin{document}
\maketitle

\begin{abstract}
Neural networks are one of the most popular approaches for many natural language processing tasks such as sentiment analysis. They often outperform traditional machine learning models and achieve the state-of-art results on most tasks.  However, many existing deep learning models are complex, difficult to train, and provide limited improvement over simpler methods. We propose a simple, robust and powerful model for sentiment classification. This model outperforms many deep learning models and achieves comparable results to other deep learning models with complex architectures on sentiment analysis datasets. We publish the code\footnote{https://bitbucket.org/taoyds/nbsvm\_pos} online.

\end{abstract}

\section{Introduction}

Deep learning models have been applied to tasks in natural language processing in recent years and have achieved very good performance.  In particular, for many sentiment analysis tasks, recurrent neural networks with many hidden layers are the current state-of-the-art.  These models perform well because they can account for the context in which words appear and are able to model long-range dependencies which may modify the sentiment of a statement within a sentence.

However, these models are complex with many parameters.  As a result, the training time is computationally intensive and may only be accessible to organizations with significant computing resources.  Furthermore, due to the many parameters in a neural network, it is easy to overfit to training data without expert knowledge of how to tune parameters.

On the other hand, one of the best performing linear models is the Naive Bayes Support Vector Machine (NBSVM) \cite{wang-manning:2012:ACL2012short}, which is a bag-of-words (BoW) model using only n-gram features.  Among traditional machine learning methods this model performs well but compared to many deep learning models it has significantly lower results.  In other work, word embeddings have been shown to be a useful resource for many natural language processing tasks.  Some research has shown how to use word embeddings in traditional models. However, there have not been many models using the combination of word embeddings and bag-of-words features.  We provide a simple method for using word embeddings in conjunction with NBSVM that outperforms more sophisticated methods.

We show that the performance of a linear bag-of-words model increases by naively combining n-grams and word embeddings.  We obtain additional improvements by grouping word embeddings according to part-of-speech (POS) tags. Finally, we discuss other complex ways of incorporating word embeddings that do not improve performance.

Section \ref{sec:rel} will describe previous work in sentiment analysis.  Sections \ref{sec:data} and \ref{sec:methods} describe the datasets used for experiments and an explanation of the methodology.  Then we present results in Section \ref{sec:results} showing near state-of-the-art performance in many tasks and outperforming several more complicated models.  Finally, we discuss our findings and future directions for this work in Section \ref{sec:conclusion}.

\section{Related Work}
\label{sec:rel}
Sentiment analysis has a rich literature of machine learning models.
Within the deep learning framework, recurrent, recursive, and convolutional networks have all been used.  
The current state of the art on many sentiment tasks is the AdaSent model \cite{DBLP:conf/ijcai/ZhaoLP15}.  This model
forms a hierarchy of representations from words to phrases and then to sentences through a recursive gated local composition of adjacent segments using recurrent and recursive neural networks before, which leads into a gating network  to form an ensemble.  One advantage of this model is that the binary parse trees for the recursive network are learned adaptively, rather than using pre-trained parses.
Using syntactic trees for recursive neural networks was first shown to perform well on sentiment analysis tasks, given labels at every node in the tree \cite{socher-EtAl:2013:EMNLP}.  
Dai and Le \shortcite{Dai:2015:SSL:2969442.2969583} used semi-supervised Sequence Learning with LSTM recurrent networks.

In the one of the first uses of convolutional neural networks for sentiment analysis,  researchers used static and non-static channels for fixed and dynamic embeddings (allowing the embeddings to vary during training time) \cite{kim:2014:EMNLP2014}.  
In other work \cite{zhou-EtAl:2016:COLING2}, researchers use a bi-directional LSTM over sentences, then treat the resulting matrix as an image and use 2-dimensional CNNs followed by max pooling (BLSTM-2DPooling).  
Another recent model, Dependency Sensitive Convolutional Neural Networks (DSCNN) hierarchically builds text representations from root-level labels using LSTMs and CNNs \cite{zhang-lee-radev:2016:N16-1}.

In the area of target-dependendent sentiment analysis, initial work used an SVM with dependency parse features \cite{jiang-EtAl:2011:ACL-HLT2011}.  Additional work also used LSTMs modified to be target-specific \cite{dong-EtAl:2014:P14-2} \cite{DBLP:conf/ijcai/VoZ15} \cite{tang-qin-liu:2015:EMNLP}.

In other work not using neural networks, NBSVM adjusts the binary counts of words in a bag-of-words approach by weighting each word according to the ratio of its counts in positive and negative documents \cite{wang-manning:2012:ACL2012short}.   More recently,
some experiments have been done to combine bag-of-words vectors with word embeddings, using the most frequent 30,000 uni-grams and bi-grams and concatenating them with averaged word embeddings \cite{DBLP:journals/corr/ZhangW15b}.  Other research involved learning weights for embeddings \cite{li-EtAl:2016:COLING5} using a Naive Bayes approach.

\section{Data}
\label{sec:data}
We test our model on several sentiment analysis datasets: the Movie Review dataset (MR), the Customer Review corpus (CR),  the MPQA Opinion corpus (MPQA), the subjectivity dataset (Subj), reviews from the Internet
Movie Database (IMDB), and a Twitter sentiment dataset (TTwt).  The MR dataset was created from short movie reviews, with one sentence per review \cite{Pang+Lee:05a}.  Another dataset (CR) was created by crawling customer reviews from the Web for 5 technology products \cite{Hu:2004:MSC:1014052.1014073} and 9 other technology products, combined following
\cite{nakagawa-inui-kurohashi:2010:NAACLHLT}.  For MPQA, we focus on the opinion polarity subtask \cite{Wiebe05annotatingexpressions}.  The goal of the subjectivity dataset (Subj) is to distinguish between subjective
reviews and objective plot summaries \cite{pang-lee:2004:ACL}.  The IMDB corpus consists of full-length reviews
\cite{maas-EtAl:2011:ACL-HLT2011} and the Twitter dataset (TTwt) \cite{jiang-EtAl:2011:ACL-HLT2011} contains sentiment dependent on a particular target.
Detailed statistics of the datasets are shown in Table 1. 

\begin{table*}[t]
\centering
\begin{tabular}{|l|l|l|l|l|l|l|l|}
\hline
Data & Classes & Length & (${\cal N+}$, ${\cal N-}$, ${\cal N}$) & Vocabulary Size & Testing  \\
\hline
MR & 2 & 20 & (5331, 5331, -) & 21000 & CV  \\
\hline
CR & 2 & 20 & (2406, 1366, -) & 5713 & CV  \\
\hline
MPQA & 2 & 3 & (3316, 7308, -) & 6299 & CV  \\
\hline
Subj & 2 & 24 & (5000, 5000, -) & 24000 & CV  \\
\hline
IMDB & 2 & 231 & (25000, 25000, -) & 392000 & N  \\
\hline
TTwt & 3 & ~13 & (1562, 1562, 3124) & ~10000 & N  \\
\hline
\end{tabular}
\caption{\footnotesize Dataset statistics. Length: Average number of words. (${\cal N+}$, ${\cal N-}$, ${\cal N}$): number of
positive, negative and neutral examples.Testing: CV for reporting results of cross-validation, N for reporting results on heldout data.}
\label{table:data}
\end{table*}

\section{Methods}
\label{sec:methods}
Our model is the Naive Bayes Logistic Regression with word embedding features (NBLR + POSwemb) model, which is an extension of NBSVM. To leverage sparse and dense feature combinations, our NBLR + POSwemb model uses the following features:

\begin{itemize}
    \item {\bf Sentiment and negation word indicators:} For each sentence, we first append positive or negative sentiment indicator tokens at the end of each sentence if it includes some words in the MPQA \cite{wilson-wiebe-hoffmann:2005:HLTEMNLP} and Liu \cite{Hu:2004:MSC:1014052.1014073} sentiment lexicons. Also, we apply the same step if the sentence contains negation words and adversatives.
    \item {\bf Log-count ratios for multiclasses:} After adding sentiment and negation word indicator to the sentence $x$, we compute log-ratio vectors for multiclasses. The count vectors for documents with label $l$ are $p_l = \alpha + \sum_{i:y^{(i)}=l} f(i)$ and for documents with other labels are $q_l = \alpha + \sum_{i:y^{(i)}\neq l} f(i)$ where $\alpha$ is a smoothing parameter (here we set $\alpha$ = 1).
    The log-count ratio for class $l$ is then:
    $$r_l = \log  \left( \frac{p_l/||p_l||_1}{q_l/||q_l||_1} \right) $$ 
    $f(i)$ is a count vector for training case $i$ with label $y^{(i)} \in \{-1, 0, 1\}$. In this work, we concatenate the log-ratio vector of each class to obtain the final sparse vector $r(x) =r_{1} \circ r_{0} \circ r_{-1}$.
    \item {\bf Averaged and part-of-speech (POS) word embedding features:} For a sentence $x$, first, we compute the averaged word embedding of all words $v_{avg}$. Then we POS tag all words in the sentence and group them into the set $P$ = \{NOUN, VERB, ADJECTIVE\} according to their POS tags.
    For each of the MR, CR, MPQA, Subj, and IMDB datasets, we use the NLTK part-of-speech tagger\footnote{http://www.nltk.org/}. For the TTwt dataset, we choose CMU Ark Tweet NLP tagging tool\footnote{https://github.com/brendano/ark-tweet-nlp}. The part-of-speech (POS) word embedding vector $v_p$ by averaging the word vectors of words which are tagged as $p$: 
    \[
    v_p(x) = \frac{\sum_{i=1}^{|x|} \mathbb{I}(pos(x_i) \in p) \times  x^i_{ce}}{\sum_{i=1}^{|x|}  \mathbb{I}(pos(x_i) \in p) }
    \]
    where $\mathbb{I}$ is the indicator feature and $x^i_{ce} \in \mathbb{R}^{d_{ce}}$ is the pre-trained word vector for the word $x_i$. The final dense vector $v(x) \in \mathbb{R}^{|P+1|\cdot d_{ce}}$ is the concatenation all the vectors $v_p(x)$ and $v_{avg}$ \footnote{For the target dependent Twitter sentiment analysis task (TTwt), we also add an additional averaged word embedding vector for the target}.
\end{itemize}

Finally, we combine the sparse and dense feature vectors to generate the final vector input of each sentence $r(x) \circ v(x)$ to a logistic regression classifier. 

We use the same pre-processing steps in (Kim, 2014) and the Google word2vec pre-trained word embeddings to compute the average vectors of words in each group \footnote{https://code.google.com/archive/p/word2vec/}. 

\begin{table*}[t]
\centering
\begin{tabular}{|l|l|l|l|l|l|l|}
\hline
Data & MR & CR & MPQA & Subj & TTwt & IMDB \\
\hline
NBSVM (Wang and Manning, 2012) & 79.4 & 81.8 & 86.3 & 93.2 & 65.6 & 91.2\\
\hline
NBLR + POSwemb & \textbf{81.6} & \textbf{84.0} & \textbf{89.9} & 93.3 & \underline{\textbf{69.9}} & 91.8 \\
\hline
\hline
Li et al., 2016 & 79.5 & 81.1 & 82.1 & 92.8 & -- & \underline{\textbf{93.0}}\\
\hline
DSCNN(Zhang et al., 2016) & 81.5 & -- & -- & 93.2 & -- & 90.2\\
\hline
BLSTM-2DPooling(Zhou et al., 2016) & 81.5 & -- & -- & \textbf{93.7} & -- & --\\
\hline
SA-LSTM (Dai and Le, 2015) & 80.7 & -- & -- & -- & -- & \textbf{92.8}\\
\hline
AdaSent (Zhao et at., 2015) & \underline{\textbf{83.1}} & \underline{\textbf{86.3}} & \underline{\textbf{93.3}} & \underline{\textbf{95.5}} & -- & --\\
\hline
CNN non-static (Kim, 2014) & 81.2 & \textbf{84.0} & 89.6 & 93.4 & -- & --\\
\hline
bowwvSVM (Zhang and Wallace, 2016) & 79.67 & 81.3 & 89.7 & 91.7 & -- & --\\
\hline
SVM-dep (Jiang et al., 2011) & -- & -- & -- & -- & 63.3 & --\\
\hline
AdaRNN comb (Dong et al., 2014) & -- & -- & -- & -- & 65.9 & --\\
\hline
Targ-dept+(Vo Zhang et al., 2015) & -- & -- & -- & -- & \underline{\textbf{69.9}} & --\\
\hline
TC-LSTM (Tang et al., 2015) & -- & -- & -- & -- & \textbf{69.5} & --\\
\hline
\end{tabular}
\label{table:results}
\caption{\footnotesize Results Compared to other Models. The bold and underlined, and bold-only numbers are the best and second best results respectively. Our NBLR + POSwemb performs the best or second on several datasets.}
\end{table*}

\section{Results and Discussion}
\label{sec:results}
Results of our experiments are shown in Table 4 
To report the results, we use either 10-fold cross-validation or train/test split depending on what is standard for the dataset. The testing column of Table 1 
specifies which method is used.  All results are reported in terms of accuracy, except for TTwt, where we report macro F-measure.

Compared to NBSVM, the performance of the NBLR + POSwemb model increases by 2-3\%. Our simple model outperforms all other recent complex neutral network models except AdaSent \cite{DBLP:conf/ijcai/ZhaoLP15} and achieves state-of-the-art performance on most of benchmarks. 

\section{Conclusion}
\label{sec:conclusion}
In this paper we have presented a new powerful model built on top of NBSVM. 
Using linear models with features derived from n-grams and embeddings, we are able to obtain near state-of-the-art results.
These simple models provide a straightforward way for practitioners to create models and run experiments on new tasks.

In the future, we plan to experiment more with ways of combining word embeddings.  Part-of-speech tags provide useful indicators of sentiment because separating nouns, verbs, and adjectives captures some high-level information about
what the sentence is about in terms of subjects and objects.
We may find that other word-level tags or constituent and dependency parse tree information is useful as well.

Models with Naive Bayes weighting, indicators, and word embeddings achieve near state-of-the-art scores on many sentiment benchmark datasets. Unlike other state-of-the-art models, our model is simple and fast to train compared to complex deep learning architectures, using only transformed uni/bi-grams and word embedding features.

\bibliographystyle{acl}
\bibliography{refs}

\begin{thebibliography}{}

\bibitem[\protect\citename{Dai and Le}2015]{Dai:2015:SSL:2969442.2969583}
Andrew~M. Dai and Quoc~V. Le.
\newblock 2015.
\newblock Semi-supervised sequence learning.
\newblock In {\em Proceedings of the 28th International Conference on Neural
  Information Processing Systems}, NIPS'15, pages 3079--3087, Cambridge, MA,
  USA. MIT Press.

\bibitem[\protect\citename{Dong \bgroup et al.\egroup
  }2014]{dong-EtAl:2014:P14-2}
Li~Dong, Furu Wei, Chuanqi Tan, Duyu Tang, Ming Zhou, and Ke~Xu.
\newblock 2014.
\newblock Adaptive recursive neural network for target-dependent twitter
  sentiment classification.
\newblock In {\em Proceedings of the 52nd Annual Meeting of the Association for
  Computational Linguistics (Volume 2: Short Papers)}, pages 49--54, Baltimore,
  Maryland, June. Association for Computational Linguistics.

\bibitem[\protect\citename{Hu and Liu}2004]{Hu:2004:MSC:1014052.1014073}
Minqing Hu and Bing Liu.
\newblock 2004.
\newblock Mining and summarizing customer reviews.
\newblock In {\em Proceedings of the Tenth ACM SIGKDD International Conference
  on Knowledge Discovery and Data Mining}, KDD '04, pages 168--177, New York,
  NY, USA. ACM.

\bibitem[\protect\citename{Jiang \bgroup et al.\egroup
  }2011]{jiang-EtAl:2011:ACL-HLT2011}
Long Jiang, Mo~Yu, Ming Zhou, Xiaohua Liu, and Tiejun Zhao.
\newblock 2011.
\newblock Target-dependent twitter sentiment classification.
\newblock In {\em Proceedings of the 49th Annual Meeting of the Association for
  Computational Linguistics: Human Language Technologies}, pages 151--160,
  Portland, Oregon, USA, June. Association for Computational Linguistics.

\bibitem[\protect\citename{Kim}2014]{kim:2014:EMNLP2014}
Yoon Kim.
\newblock 2014.
\newblock Convolutional neural networks for sentence classification.
\newblock In {\em Proceedings of the 2014 Conference on Empirical Methods in
  Natural Language Processing (EMNLP)}, pages 1746--1751, Doha, Qatar, October.
  Association for Computational Linguistics.

\bibitem[\protect\citename{Li \bgroup et al.\egroup
  }2016]{li-EtAl:2016:COLING5}
Bofang Li, Zhe Zhao, Tao Liu, Puwei Wang, and Xiaoyong Du.
\newblock 2016.
\newblock Weighted neural bag-of-n-grams model: New baselines for text
  classification.
\newblock In {\em Proceedings of COLING 2016, the 26th International Conference
  on Computational Linguistics: Technical Papers}, pages 1591--1600, Osaka,
  Japan, December. The COLING 2016 Organizing Committee.

\bibitem[\protect\citename{Maas \bgroup et al.\egroup
  }2011]{maas-EtAl:2011:ACL-HLT2011}
Andrew~L. Maas, Raymond~E. Daly, Peter~T. Pham, Dan Huang, Andrew~Y. Ng, and
  Christopher Potts.
\newblock 2011.
\newblock Learning word vectors for sentiment analysis.
\newblock In {\em Proceedings of the 49th Annual Meeting of the Association for
  Computational Linguistics: Human Language Technologies}, pages 142--150,
  Portland, Oregon, USA, June. Association for Computational Linguistics.

\bibitem[\protect\citename{Nakagawa \bgroup et al.\egroup
  }2010]{nakagawa-inui-kurohashi:2010:NAACLHLT}
Tetsuji Nakagawa, Kentaro Inui, and Sadao Kurohashi.
\newblock 2010.
\newblock Dependency tree-based sentiment classification using crfs with hidden
  variables.
\newblock In {\em Human Language Technologies: The 2010 Annual Conference of
  the North American Chapter of the Association for Computational Linguistics},
  pages 786--794, Los Angeles, California, June. Association for Computational
  Linguistics.

\bibitem[\protect\citename{Pang and Lee}2004]{pang-lee:2004:ACL}
Bo~Pang and Lillian Lee.
\newblock 2004.
\newblock A sentimental education: Sentiment analysis using subjectivity
  summarization based on minimum cuts.
\newblock In {\em Proceedings of the 42nd Meeting of the Association for
  Computational Linguistics (ACL'04), Main Volume}, pages 271--278, Barcelona,
  Spain, July.

\bibitem[\protect\citename{Pang and Lee}2005]{Pang+Lee:05a}
Bo~Pang and Lillian Lee.
\newblock 2005.
\newblock Seeing stars: Exploiting class relationships for sentiment
  categorization with respect to rating scales.
\newblock In {\em Proceedings of ACL}, pages 115--124.

\bibitem[\protect\citename{Socher \bgroup et al.\egroup
  }2013]{socher-EtAl:2013:EMNLP}
Richard Socher, Alex Perelygin, Jean Wu, Jason Chuang, Christopher~D. Manning,
  Andrew Ng, and Christopher Potts.
\newblock 2013.
\newblock Recursive deep models for semantic compositionality over a sentiment
  treebank.
\newblock In {\em Proceedings of the 2013 Conference on Empirical Methods in
  Natural Language Processing}, pages 1631--1642, Seattle, Washington, USA,
  October. Association for Computational Linguistics.

\bibitem[\protect\citename{Tang \bgroup et al.\egroup
  }2015]{tang-qin-liu:2015:EMNLP}
Duyu Tang, Bing Qin, and Ting Liu.
\newblock 2015.
\newblock Document modeling with gated recurrent neural network for sentiment
  classification.
\newblock In {\em Proceedings of the 2015 Conference on Empirical Methods in
  Natural Language Processing}, pages 1422--1432, Lisbon, Portugal, September.
  Association for Computational Linguistics.

\bibitem[\protect\citename{Vo and Zhang}2015]{DBLP:conf/ijcai/VoZ15}
Duy{-}Tin Vo and Yue Zhang.
\newblock 2015.
\newblock Target-dependent twitter sentiment classification with rich automatic
  features.
\newblock In {\em Proceedings of the Twenty-Fourth International Joint
  Conference on Artificial Intelligence, {IJCAI} 2015, Buenos Aires, Argentina,
  July 25-31, 2015}, pages 1347--1353.

\bibitem[\protect\citename{Wang and
  Manning}2012]{wang-manning:2012:ACL2012short}
Sida Wang and Christopher Manning.
\newblock 2012.
\newblock Baselines and bigrams: Simple, good sentiment and topic
  classification.
\newblock In {\em Proceedings of the 50th Annual Meeting of the Association for
  Computational Linguistics (Volume 2: Short Papers)}, pages 90--94, Jeju
  Island, Korea, July. Association for Computational Linguistics.

\bibitem[\protect\citename{Wiebe and Cardie}2005]{Wiebe05annotatingexpressions}
Janyce Wiebe and Claire Cardie.
\newblock 2005.
\newblock Annotating expressions of opinions and emotions in language. language
  resources and evaluation.
\newblock In {\em Language Resources and Evaluation (formerly Computers and the
  Humanities}, page 2005.

\bibitem[\protect\citename{Wilson \bgroup et al.\egroup
  }2005]{wilson-wiebe-hoffmann:2005:HLTEMNLP}
Theresa Wilson, Janyce Wiebe, and Paul Hoffmann.
\newblock 2005.
\newblock Recognizing contextual polarity in phrase-level sentiment analysis.
\newblock In {\em Proceedings of Human Language Technology Conference and
  Conference on Empirical Methods in Natural Language Processing}, pages
  347--354, Vancouver, British Columbia, Canada, October. Association for
  Computational Linguistics.

\bibitem[\protect\citename{Zhang and
  Wallace}2015]{DBLP:journals/corr/ZhangW15b}
Ye~Zhang and Byron~C. Wallace.
\newblock 2015.
\newblock A sensitivity analysis of (and practitioners' guide to) convolutional
  neural networks for sentence classification.
\newblock {\em CoRR}, abs/1510.03820.

\bibitem[\protect\citename{Zhang \bgroup et al.\egroup
  }2016]{zhang-lee-radev:2016:N16-1}
Rui Zhang, Honglak Lee, and Dragomir~R. Radev.
\newblock 2016.
\newblock Dependency sensitive convolutional neural networks for modeling
  sentences and documents.
\newblock In {\em Proceedings of the 2016 Conference of the North American
  Chapter of the Association for Computational Linguistics: Human Language
  Technologies}, pages 1512--1521, San Diego, California, June. Association for
  Computational Linguistics.

\bibitem[\protect\citename{Zhao \bgroup et al.\egroup
  }2015]{DBLP:conf/ijcai/ZhaoLP15}
Han Zhao, Zhengdong Lu, and Pascal Poupart.
\newblock 2015.
\newblock Self-adaptive hierarchical sentence model.
\newblock In {\em Proceedings of the Twenty-Fourth International Joint
  Conference on Artificial Intelligence, {IJCAI} 2015, Buenos Aires, Argentina,
  July 25-31, 2015}, pages 4069--4076.

\bibitem[\protect\citename{Zhou \bgroup et al.\egroup
  }2016]{zhou-EtAl:2016:COLING2}
Peng Zhou, Zhenyu Qi, Suncong Zheng, Jiaming Xu, Hongyun Bao, and Bo~Xu.
\newblock 2016.
\newblock Text classification improved by integrating bidirectional lstm with
  two-dimensional max pooling.
\newblock In {\em Proceedings of COLING 2016, the 26th International Conference
  on Computational Linguistics: Technical Papers}, pages 3485--3495, Osaka,
  Japan, December. The COLING 2016 Organizing Committee.

\end{thebibliography}

\end{document}